\documentclass[orivec,runningheads]{llncs}
\usepackage{amsmath,amssymb} 
\usepackage{color}

\usepackage{cite}

\usepackage{url}

\usepackage{rotating}
\usepackage{multirow}
\usepackage[lofdepth,lotdepth,caption=false]{subfig}
\definecolor{Gray}{gray}{0.8}
\usepackage{hhline}
\usepackage{color}
\usepackage{colortbl}

\usepackage{MnSymbol} 

\usepackage{footmisc} 

\newcommand{\ie}{i.e. }
\newcommand{\etal}{et al.}

\newcommand{\vertex}{{\bf v}}
\newcommand{\xClPoint}{{X}}
\newcommand{\ob}{{\bf o}}
\newcommand{\normal}{{\bf n}}
\newcommand{\ld}{{\bf d}}
\newcommand{\lm}{{\bf m}}
\newcommand{\para}{{\theta}}
\newcommand{\argmin}{\mathop{\mathgroup\symoperators argmin}}
\newcommand{\pluecker}{Pl{\"u}cker }

\begin{document}
\pagestyle{headings}
\mainmatter

\def\DAGM12SubNumber{151}  

\title{Capturing Hand Motion with an RGB-D Sensor, Fusing a Generative Model with Salient Points}

\titlerunning{Capturing Hand Motion with an RGB-D Sensor}
\authorrunning{D. Tzionas \and A. Srikantha \and P. Aponte \and J. Gall}
\author{Dimitrios Tzionas\inst{1}\textsuperscript{,}\inst{2} \and Abhilash Srikantha\inst{1}\textsuperscript{,}\inst{2} \and Pablo Aponte\inst{2} \and Juergen Gall\inst{2}}
\urldef{\mailsa}\path|dimitris.tzionas@tue.mpg.de|
\urldef{\mailsb}\path|abhilash.srikantha@tue.mpg.de|
\urldef{\mailsc}\path|aponte@iai.uni-bonn.de|
\urldef{\mailsd}\path|gall@iai.uni-bonn.de|
\institute{Perceiving Systems Department, MPI for Intelligent Systems, Germany\\ \mailsa, \mailsb \\ \and Computer Vision Group, University of Bonn, Germany\\  \mailsc, \mailsd}

\maketitle

\setcounter{footnote}{0}

\begin{abstract}
Hand motion capture has been an active research topic, following the success of full-body pose tracking. 
Despite similarities, hand tracking proves to be more challenging, characterized by a higher dimensionality, severe occlusions and self-similarity between fingers. 
For this reason, most approaches rely on strong assumptions, like hands in isolation or expensive multi-camera systems, that limit practical use. 
In this work, we propose a framework for hand tracking that can capture the motion of two interacting hands using only a single, inexpensive RGB-D camera. 
Our approach combines a generative model with collision detection and discriminatively learned salient points. 
We quantitatively evaluate our approach on 14 new sequences with challenging interactions.

\end{abstract}

\section{Introduction}\label{sec:introduction}
\vspace{-2mm}

	Human body tracking has been a popular field of research during the past decades \cite{Review_Moeslund_2006}, recently gaining more popularity due to the ubiquity of RGB-D sensors. 
	Hand motion capture, a special instance of it, has enjoyed much research interest \cite{Review_Erol_HandPose} due to its numerous applications including, but not limited to, 
	computer graphics, human-computer-interaction and robotics. 
	
	Despite similarities, robust techniques \cite{kinect_Paper} for full-body tracking 
	are insufficient for hand motion capture, as the latter is more complicated on numerous fronts. 
	Hands are characterized by more degrees of freedom, formulating a higher dimensional optimization problem. 
	Severe occlusions are a usual phenomenon, being either self-occlusions or occlusions from another hand or object. 
	Similarity in shape and appearance causes ambiguities for the differentiation between fingers and hands.
	Fast motion and lower resolution of hands in images constitute further complicating factors. 

	Despite theses challenges, there has been substantial progress in hand motion capture in recent years.  
	Ballan \etal~\cite{LucaHands} have presented a system that successfully captures the motion of two hands strongly interacting with each other and an additional object. Although the approach achieves remarkable accuracy, it is based on an expensive and elaborate multi-camera system. 
	On the other hand, Oikonomidis \etal~\cite{OikonomidisBMVC,Oikonomidis_1hand_object,Oikonomidis_2hands} have presented a real time hand tracker using just a single off-the-shelf RGB-D camera.
	Despite their success under challenging scenarios, the exhibited accuracy is not as precise as \cite{LucaHands}.
	
	Our approach for tracking the pose of two strongly interacting hands is inspired by Ballan \etal~\cite{LucaHands} 
	and combines a generative model with an occlusion handling method and a discriminatively trained detector for salient points. While~\cite{LucaHands} relies on an expensive capture setup with 8 synchronized and calibrated RGB cameras recording FullHD footage at 50 fps, we propose an approach that captures hand motion of two interacting hands using a cheap RGB-D camera recording VGA resolution at 30 fps. 
	We evaluate our approach on 14 annotated sequences\footnote{\label{projectWebsite}The annotated dataset sequences and the supplementary material are available at \url{http://files.is.tue.mpg.de/dtzionas/GCPR_2014.html}.}, which include interactions between hands. 
	We further compare our method to the single hand tracker~\cite{OikonomidisBMVC} on sequences with one hand, showing that our approach estimates the hand pose with higher accuracy than~\cite{OikonomidisBMVC}.

	\begin{figure}[t]
	\captionsetup[subfigure]{labelformat=empty}
	\centering
		\subfloat[subfigure 1 CompFORTH][]{
			\includegraphics[width=0.15 \textwidth]{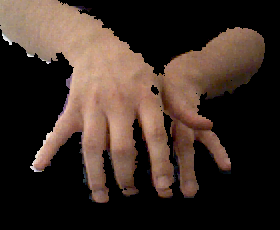}
			\label{fig:muResult_102_rgbd}
		}
		\subfloat[subfigure 2 CompFORTH][]{
			\includegraphics[width=0.15 \textwidth]{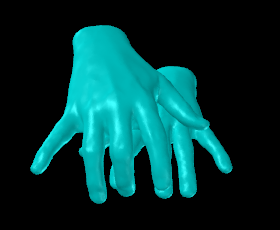}
			\label{fig:muResult_102_synth}
		}
		\subfloat[subfigure 3 CompFORTH][]{
			\includegraphics[width=0.15 \textwidth]{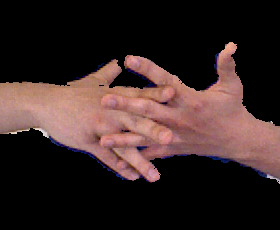}
			\label{fig:muResult_109_rgbd}
		}
		\subfloat[subfigure 4 CompFORTH][]{
			\includegraphics[width=0.15 \textwidth]{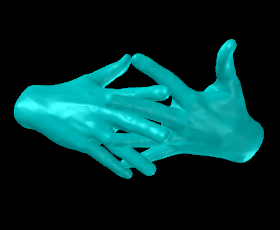}
			\label{fig:muResult_109_synth}
		}
		\subfloat[subfigure 5 CompFORTH][]{
			\includegraphics[width=0.15 \textwidth]{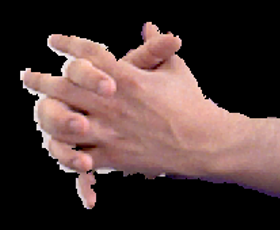}
			\label{fig:muResult_122_rgbd}
		}
		\subfloat[subfigure 6 CompFORTH][]{
			\includegraphics[width=0.15 \textwidth]{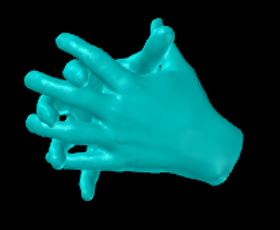}
			\label{fig:muResult_122_synth}
		}
	\vspace{-6mm}
	\caption{	Qualitative results of our pipeline. Each pair shows the aligned RGB and depth input maps after depth thresholding, along with the pose estimate output
	}
	\label{fig:resultsQualitative}
	\end{figure}
	
\section{Related Work}\label{sec:relatedWork}
\vspace{-2mm}
	
	The study of hand motion tracking has its roots in the $90$s \cite{Kanade94,Kanade95}. 
	Although the problem can be simplified by means of data-gloves~\cite{Kragic_dataglove}, color-gloves \cite{glove_MIT}, markers \cite{colorGlove} or wearable sensors \cite{Oikonomidis_MSR_Digits}, 
	the ideal solution pursued is the unintrusive, marker-less capture of hand motion.	
	Even until recently the study was mainly confined to the case of a single isolated hand \cite{AthitsosCluttered2003,HoggHand96,Isard2000,OikonomidisBMVC,srinath_iccv2013,Cipolla_ModelBased,Cipolla_TORR}. 
	However, in pursuit of more realistic scenarios, research effort was directed towards the case of a hand interacting with an object~\cite{Hamer_Hand_Manipulating,Hamer_ObjectPrior,Oikonomidis_1hand_object}, 
	two hands interacting with each other \cite{LucaHands,Oikonomidis_2hands} and with an additional object \cite{LucaHands}.
	Multiple objects can be tracked by means of hand tracking and physical forces modeling \cite{kyriazis2013}.

	An analytical review of the field can be found in the work of Erol \etal~\cite{Review_Erol_HandPose}.		
	In this work a taxonomy is presented, separating the methods met in the literature in two main categories, namely \emph{model-based} and \emph{appearance-based}.

	Generative-model approaches \cite{HoggHand96,PonsModelBased,Cipolla_ModelBased} are based on an explicit model used to generate pose hypotheses, which are evaluated against the observed data. 
	The evaluation is based on an objective function which implicitly measures the likelihood
	by computing the discrepancy between the pose estimate (hypothesis) and the observed data in terms of an error metric.
	To keep the problem tractable, 
	each iteration is initialized by the pose estimate of the previous step, relying thus heavily on temporal continuity and being prone to accumulative error.
	The objective function is evaluated in the high-dimensional, continuous parameter space.

	Discriminative methods learn a direct mapping from the observed image features to the discrete \cite{AthitsosCluttered2003,Romero09,JavierHandsInAction} or continuous \cite{Murray_handRegression2006,KeskinECCV12,Athitsos_HandSpecializedMappings2001,kinect_Paper} target parameter space. 
	Most methods operate on a single frame \cite{AthitsosCluttered2003,Murray_handRegression2006,Athitsos_HandSpecializedMappings2001,kinect_Paper}, being thus immune to pose-drifting due to error accumulation.
	Generalization in terms of capturing illumination, articulation and view-point variation, can be realized only through adequate representative training data. 
	Acquisition and annotation of realistic training data 
	is though a cumbersome and costly procedure.
	For this reason most approaches rely on synthetic rendered data \cite{KeskinECCV12,JavierHandsInAction,kinect_Paper} that has inherent ground-truth. 
	However, the discrepancy between realistic and synthetic data is an important limiting factor, while special care is needed to avoid over-fitting to the training set. 
	The accuracy of discriminative methods heavily depends on the invariance, repeatability and discriminative properties of the features employed 
	and is lower in comparison to generative methods.

	A discriminative method can effectively complement a generative method, either in terms of initialization or recovery, 
	driving the optimization framework away from local minima in the search space and aiding convergence to the global minimum. 
	Sridhar \etal~\cite{srinath_iccv2013} combine in a real time system a Sums-of-Gaussians generative model with a discriminatively trained fingertip detector in depth images using a linear SVM classifier.
	Ballan \etal \cite{LucaHands} present an accurate offline tracker that combines in a single framework a generative model 
	with a salient-point (finger-nail) Hough-forest \cite{juergen_Hough} detector in color images. Both approaches~\cite{LucaHands,srinath_iccv2013}, however, require an expensive multi-camera hardware setup.

\section{Tracking Method}\label{sec:trackingMethod}
\vspace{-2mm}
	
		\subsection{Hand Model}\label{sec:handModel}\label{sec:poseParameterization}
 
		We resort to the \emph{Linear Blend Skinning} (LBS) model~\cite{LBS_PoseSpace}, consisting of a triangular mesh, an underlying kinematic skeleton and a set of skinning weights. 
		In our experiments, a 
		triangular mesh of a pair of hands
		was obtained by a commercial 3D scanning solution and the skeleton structure was manually defined. 
		Additional details are provided in the supplementary material\footref{projectWebsite}. 
		The skinning weight $\alpha_{\vertex,j}$ defines the influence of bone $j$ on 3D vertex $\vertex$, where $\sum_j \alpha_{\vertex,j} = 1$.
		The deformation of the mesh is driven by the underlying skeleton with pose parameter vector $\para$ through the skinning weights and is expressed by the LBS operator: 
		\begin{equation}\label{eq:lbs}
		\vertex(\para) = \sum_j \alpha_{\vertex,j} T_j(\para) T_j(0)^{-1} \vertex(0)
		\end{equation}
		where $T_j(0)$ and $\vertex(0)$ are the bone transformations and vertex positions at the known rigging pose.
		The skinning weights are computed using~\cite{pinocchio}. 
	
		The global rigid motion is represented by a twist~\cite{Malik_Twist,Murray_MathInRob,PonsModelBased} in the special Euclidean group $SE(3)$. 
		The articulation of the skeleton is expressed by a kinematic chain of rigid components. 
		For the sake of simplicity, joints with more than 1 degree of freedom (DoF) are modeled by a combination of revolute joints. Using the exponential map operator, the transformation of a bone $T_j(\para)$ with $k$ DoF is therefore given by $T_j(\para) = \prod_{i<j}T_i(\para)\prod_{k_j} \exp(\theta_{k_j} \hat{\xi}_{k_j})$, where $\para$ is the parameterization of the full pose, $\theta_{k_j} \hat{\xi}_{k_j}$ is the twist representation of a single revolute joint, and $T_{i<j}$ denotes all previous bones in the kinematic chain.   
		
		In our experiments, a single hand consists of 31 revolute joints, \ie 37 DoF. 
		Thus, for sequences with two interacting hands we have to estimate all 74 DoF.
		
		Anatomically inspired joint-angle limits \cite{AnatomicalModel} constrain the solution space to the subspace of physically plausible poses as in \cite{LucaHands}. 
		
\vspace{-2mm}
\subsection{Optimization}\label{sec:optimization}
	
	Our objective function for pose estimation consists of four terms:
	\begin{align}\label{eq:obj}
	\begin{split}
	E(\para, D) = \quad&E_{model \rightarrow data}(\para,D) + E_{data \rightarrow model}(\para,D) + \\  
		       &E_{salient}(\para,D) + \gamma_c E_{collision}(\para) 
	\end{split}
	\end{align}
	where $\para$ are the pose parameters of the two hands and $D$ is the current preprocessed depth image. The first two terms minimize the alignment error of the transformed mesh and the depth data. The alignment error is measured by $E_{model \rightarrow data}$, which measures how well the model fits the observed depth data, and $E_{data \rightarrow model}$, which measures how well the depth data is explained by the model. The last two terms are inspired by~\cite{LucaHands}. 
	$E_{salient}$ measures the consistency of the generative model with detected salient points in the image. The main purpose of the term in our framework is to recover from tracking errors of the generative model. In our scenario with a single camera of low resolution, the 3D positions of the detected points are less accurate and additional care is needed. 
	$E_{collision}$ penalizes intersections of fingers, ensuring physically plausible poses.
	
	The objective Equation \eqref{eq:obj} is minimized by local optimization as described in~\cite{PonsModelBased}. In the following, we give details for the terms of the objective function.

\vspace{-2mm}
	\subsubsection{Preprocessing:}\label{sec:Preprocessing}

	For pose estimation, we first remove irrelevant parts of the RGB-D image by thresholding the depth values and applying skin color segmentation~\cite{skinnColorGMM} on the RGB image. 
	As a result, we get a masked RGB-D image, which is denoted as $D$ in Equation \eqref{eq:obj}. The thresholding of the depth image avoids unnecessary processing like normal computation for points far away and the skin color segmentation removes occluding objects from the data.  
		
\vspace{-2mm}
	\subsubsection{Fitting the \emph{model to the data}:}\label{subsec:m2d}

	The first term in Equation \eqref{eq:obj} aims at fitting the mesh parameterized by pose parameters $\para$ to the preprocessed data $D$. 
	To this end, the depth values are converted into a 3D point cloud based on the calibration data of the sensor.	
	The point cloud is then smoothed by a bilateral filter \cite{bilateralFAST} and normals are computed \cite{normals_integralImages_Holzer}. 
	For each vertex of the model $\vertex_{i}(\para)$, with normal $\normal_{i}(\para)$, we search for the closest point $\xClPoint_i$ in the point cloud. This gives a 3D-3D correspondence for each vertex. 
	We discard the correspondence if the angle between the normals of the vertex and the closest point is larger than 45$^\circ$ or the distance between the points is larger than 10 mm.  
	We can then write the term $E_{model \rightarrow data}$ as a least squared error of \emph{point-to-point} distances:
	\begin{equation}\label{eq:errorResidual_p2p}
	E_{model \rightarrow data}(\para,D) = \sum_i \Vert \vertex_{i}(\para) - \xClPoint_i \Vert^2\;
	\end{equation}
	An alternative to the \emph{point-to-point} distance is the \emph{point-to-plane} distance, which is commonly used for 3D reconstruction~\cite{p2pl,ICP_EfficientVariants,RusinkiewiczRealTimeINHAND}:
	\begin{equation}\label{eq:errorResidual_p2pl}
	E_{model \rightarrow data}(\para,D) = \Vert \normal_{i}(\para)^T (\vertex_{i}(\para) - \xClPoint_i) \Vert^2\;
	\end{equation}
	The two distance metrics are evaluated in our experiments (Section \ref{subsubsec:experimentDistanceMetrics}). 
	In general, the \emph{point-to-plane} distance converges faster and is therefore preferred.  
	
	\subsubsection{Fitting the \emph{data to the model}:}\label{subsec:d2m}
	Only fitting the model to the data is not sufficient as we will show in our experiments. In particular, poses with self-occlusions can have a very low error since the measure only evaluates how well the visible part of the model fits the point cloud. 
	The second term $E_{data \rightarrow model}(\para,D)$ matches the data to the model to make sure that the solution is not degenerate and explains the data as well as possible. 
	Since matching the data to the model is expensive, we reduce the matching to depth discontinuities~\cite{functionalCategorization}. 
	To this end, we extract depth discontinuities from the depth map and the projected depth profile of the model using an edge detector~\cite{canny}. 
	Correspondences are again established by searching for the closest points, but now in the depth image using a 2D distance transform~\cite{DT_Felzenszwalb}. 
	Similar to $E_{model \rightarrow data}(\para,D)$, we discard correspondences with a large distance. The depth values at the depth discontinuities in $D$, 
	however, are less reliable not only due to the depth ambiguities between foreground and background, but also due to the noise of cheap sensors. 
	The depth of the point in $D$ is therefore computed as average in a local $3x3$ pixels neighborhood and the outlier distance threshold is increased to 30 mm. 
	The approximation is sufficient for discarding outliers, but insufficient for minimization. 
	For each matched point in $D$ we therefore compute the projection ray uniquely expressed as a \pluecker line~\cite{PonsModelBased,Rosenh_Plucker_IJCV,Stolfi91} 
	with direction $\ld_i$ and moment $\lm_i$ 
	and minimize the least square error between the projection ray and the vertex $\vertex_{i}(\para)$ for each correspondence:  
	\begin{equation}\label{eq:errorResidual_Pluecker}
	E_{data \rightarrow model}(\para,D) = \sum_{i} \Vert \vertex_{i}(\para) \times \ld_i - \lm_i \Vert^2\;
	\end{equation}

	\subsubsection{Collision detection}\label{subsec:collisionDetection}
		Collision detection is based on the observation that two objects cannot share the same space and
		is of high importance in case of self-penetration, inter-finger penetration or general intensive interaction.

		For detecting collisions, we use \emph{bounding volume hierarchies} (BVH) to efficiently determine collisions between meshes~\cite{collisionDeformableObjects}. 
		Having found a collision between two triangles $f_s$ and $f_t$, the amount of penetration can be computed as in~\cite{LucaHands} using a 3D distance field in the form of a cone.
		Considering the case where the vertices of $f_s$ are the \emph{intruders} and the triangle $f_t$ is the \emph{receiver} of the penetration (similarly for the opposite case), 
		the cone for computing the 3D distance field $\Psi_{f_t}$ is defined by the circumcenter of the triangle $f_t$. 
		Letting $\normal_{f_t}$ denote the normal of the triangle, $\ob_{f_t}$ the circumcenter, and $r_{f_t}$ the radius of the circumcircle, we have 
		\begin{equation}\label{eq:collision_PSI}
		\Psi_{f_t}(\vertex_s) = 
		\begin{cases} 
		~\vert ( 1-\Phi(\vertex_s) ) \Upsilon( \normal_{f_t} \cdot( \vertex_s-\ob_{f_t} ) ) \vert^2	&\text{when}\quad\Phi(\vertex_s)<1 	\\
		~0									&\text{when}\quad\Phi(\vertex_s)\ge1
		\end{cases}
		\end{equation}	
		\vspace{-4mm}
		\begin{equation}\label{eq:collision_FFF}
		\Phi(\vertex_s) = \frac{  \Vert (\vertex_s-\ob_{f_t})-(\normal_{f_t}\cdot(\vertex_s-\ob_{f_t}))\normal_{f_t} \Vert  }{  -\frac{r_{f_t}}{\sigma} (\normal_{f_t}\cdot(\vertex_s-\ob_{f_t}))+r  }
		\end{equation}
		\begin{equation}\label{eq:collision_YYY}
		\Upsilon(x) = 
		\begin{cases} 
		-x+1-\sigma										&\text{when}\quad             x \le -\sigma		\\
		-\frac{1-2\sigma}{4\sigma^2}x^2 - \frac{1}{2\sigma}x + \frac{1}{4}(3-2\sigma)		&\text{when}\quad -\sigma \le x \le +\sigma		\\
		0											&\text{when}\quad             x \ge +\sigma
		\end{cases}
		\end{equation}
		The parameter $\sigma$ defines the field of view of the cone and is fixed to 0.5 as in~\cite{LucaHands}. 
		
		For each vertex penetrating a triangle, a force can be computed that pushes the vertex back, where the direction is given by the inverse normal direction of the vertex and the strength of the force by $\Psi$.  
		Using \emph{point-to-point} distances \eqref{eq:errorResidual_p2p}, the forces are computed for the set of colliding triangles $\mathcal{C}$:    
		\begin{equation}
		E_{collision}(\para) = \sum_{(f_s(\para), f_t(\para))\in \mathcal{C}} \left\{ \sum_{\vertex_s \in f_s} \Vert -\Psi_{f_t}(\vertex_s)\normal_s  \Vert^2 + \sum_{\vertex_t \in f_t} \Vert -\Psi_{f_s}(\vertex_t)\normal_t \Vert^2 \right\} 
		\end{equation}
		Though not explicitly denoted, $f_s$ and $f_t$ depend on $\para$ and therefore also $\Psi$, $\vertex$ and $\normal$. For \emph{point-to-plane} distances \eqref{eq:errorResidual_p2pl}, the equation gets simplified since $\normal^T\normal=1$:   
		\begin{equation}
		E_{collision}(\para) = \sum_{(f_s(\para), f_t(\para))\in \mathcal{C}} \left\{ \sum_{\vertex_s \in f_s} \Vert -\Psi_{f_t}(\vertex_s)  \Vert^2 + \sum_{\vertex_t \in f_t} \Vert -\Psi_{f_s}(\vertex_t) \Vert^2 \right\} 
		\end{equation}

		This term takes part in the objective function \eqref{eq:obj} regulated by weight $\gamma_c$. An evaluation of different $\gamma_c$ values is presented in our experiments (Section \ref{subsubsec:experimentCollisionDetection}).

	\vspace{-1mm}
	\subsubsection{Salient point detection:}\label{subsec:salientPointDetector}
	Our approach is so far based on a generative model, which generally provides accurate solutions, but recovers only slowly from ambiguities and tracking errors. It has been shown in~\cite{LucaHands} that this can be compensated by integrating a discriminatively trained salient point detector into a generative model. In~\cite{LucaHands}, a finger nail detector was applied to the high resolution images and due to the multi-camera setup it could be assumed that the nails become visible in some of the cameras. 
	For low-resolution video of a single camera, finger nails cannot be reliably detected. Instead we train a fingertip detector~\cite{juergen_Hough} on raw depth data where the training data is not part of the test sequences. 
	More details are given in the supplementary material.

	Since we resort to salient points only for additional robustness, it is usually sufficient to have only sparse fingertip detections. 
	We therefore collect detections with a high confidence, choosing a threshold of $c_{thr}=3.0$ for our experiments. 
	The association between detections and fingertips of the model, as shown in Table~\ref{table:bipartiteTable}, is solved by integer programming~\cite{LucaHands,Malik_Bipartite}:
	\begin{equation}\label{eq:bipartite_MIP}
	\begin{split}
	\argmin 		\qquad 		& \sum_{s,t}e_{st}w_{st} + \lambda\sum_{s}\alpha_s w_s + \lambda\sum_{t}\beta_t	\\
	\text{subject to} 	\qquad		& \sum_{s}e_{st} + \beta_t = 1 \qquad \forall t, 					\\
						& \sum_{t}e_{st} + \alpha_s  = 1 \qquad \forall s					\\
						& e_{st}, \alpha_t, \beta_s \in \{0,1\}
	\end{split}
	\end{equation}
	The weights $w_{st}$ are given by the 3D distance between the detection $\delta_s$ and the finger of the model $\xi_t$. 
	For each finger $\xi_t$, a set of vertices are marked in the model. 
	The distance is then computed as the centroid of the visible vertices of $\xi_t$ and the centroid of the detected region $\delta_s$. 
	For the weights $w_s$, we investigate two approaches. 
	The first approach uses $w_s=1$ as in~\cite{LucaHands}. 
	The second approach takes the confidences $c_s$ of the detections into account by setting $w_s=\frac{c_s}{c_{thr}}$. 
	The weighting parameter $\lambda$ is evaluated in the experimental section (Section \ref{subsubsec:experimentSalientPointDetector}). 
	
		\begin{table}[t]
			\footnotesize 
			\begin{center}
				\caption{	The graph contains $T$ mesh fingertips $\xi_t$ and $S$ fingertip detections $\delta_s$. 
						The cost of assigning a detection $\delta_s$ to a finger $\xi_t$ is given by $w_{st}$ as shown in table (a). 
						The cost of declaring a detection as false positive is $\lambda w_s$ where $w_s$ is the detection confidence. 
						The cost of not assigning any detection to finger $\xi_t$ is given by $\lambda$. 
						The binary solution of table (b) is constrained to sum up to $1$ for each row and column
				}
				\vspace{-2mm}
				\label{table:bipartiteTable}
				\setlength{\tabcolsep}{1pt}	
				\begin{tabular}{|c|c|c|c|c|c|c|c|c|}
					\cline{4-7}\cline{9-9}
					\multicolumn{2}{c}{\multirow{2}{*}{(a)}} & \multicolumn{1}{c}{} & \multicolumn{4}{|c|}{Fingertips $\xi_t$} & \multicolumn{1}{c}{} & \multicolumn{1}{|c|}{V} \\
					\cline{4-7}\cline{9-9}
					\multicolumn{2}{c}{} & \multicolumn{1}{c}{} & \multicolumn{1}{|c|}{$\xi_1$} & $\xi_2$ & {\dots} & $\xi_T$ & \multicolumn{1}{c}{} & \multicolumn{1}{|c|}{$\alpha$} \\
					\cline{4-7}\cline{9-9}
					\noalign{\smallskip}
					\cline{1-2}\cline{4-7}\cline{9-9}
					\multirow{5}{*}{\centering\begin{turn}{90}Detections $\delta_s$\end{turn} } & {	\textit{$\delta_1$}	} & {} & { $w_{11}$ } & { $w_{12}$ } & { $\dots$  } & { $w_{1T}$ } & {} & { $\lambda w_1$ }                \\ \cline{2-2}\cline{4-7}\cline{9-9}
														   {} & {	\textit{$\delta_2$}	} & {} & { $w_{21}$ } & { $w_{22}$ } & { $\dots$  } & { $w_{2T}$ } & {} & { $\lambda w_2$ }                \\ \cline{2-2}\cline{4-7}\cline{9-9}
														   {} & {	\textit{$\delta_3$}	} & {} & { $w_{31}$ } & { $w_{32}$ } & { $\dots$  } & { $w_{3T}$ } & {} & { $\lambda w_3$ }                \\ \cline{2-2}\cline{4-7}\cline{9-9}
														   {} & {	         \vdots  	} & {} & { $\vdots$ } & { $\vdots$ } & { $\ddots$ } & { $\vdots$ } & {} & { $\vdots                      $ }                \\ \cline{2-2}\cline{4-7}\cline{9-9}
														   {} & {	\textit{$\delta_S$}	} & {} & { $w_{S1}$ } & { $w_{S2}$ } & { $\dots$  } & { $w_{ST}$ } & {} & { $\lambda w_S$ }                \\ 
					\cline{1-2}\cline{4-7}\cline{9-9}
					\noalign{\smallskip}
					\cline{1-2}\cline{4-7}\cline{9-9}
					\multirow{1}{*}{\centering\begin{turn}{90}V\end{turn}                     } & {        \textit{$\beta$   }	} & {} & { $\lambda$ } & { $\lambda$ } & {\dots} & { $\lambda$ } & {} & { $\infty$ } \\ 
					\cline{1-2}\cline{4-7}\cline{9-9}
				\end{tabular}
				\quad
				\begin{tabular}{|c|c|c|c|c|c|c|c|c|}
					\cline{4-7}\cline{9-9}
					\multicolumn{2}{c}{\multirow{2}{*}{(b)}} & \multicolumn{1}{c}{} & \multicolumn{4}{|c|}{Fingertips $\xi_t$} & \multicolumn{1}{c}{} & \multicolumn{1}{|c|}{V} \\
					\cline{4-7}\cline{9-9}
					\multicolumn{2}{c}{} & \multicolumn{1}{c}{} & \multicolumn{1}{|c|}{$\xi_1$} & $\xi_2$ & {\dots} & $\xi_T$ & \multicolumn{1}{c}{} & \multicolumn{1}{|c|}{$\alpha$} \\
					\cline{4-7}\cline{9-9}
					\noalign{\smallskip}
					\cline{1-2}\cline{4-7}\cline{9-9}
					\multirow{5}{*}{\centering\begin{turn}{90}Detections $\delta_s$\end{turn} } & {	\textit{$\delta_1$}	} & {} & { $e_{11}$ } & { $e_{12}$ } & { $\dots$  } & { $e_{1T}$ } & {} & { $\alpha_1$ }                \\ \cline{2-2}\cline{4-7}\cline{9-9}
														   {} & {	\textit{$\delta_2$}	} & {} & { $e_{21}$ } & { $e_{22}$ } & { $\dots$  } & { $e_{2T}$ } & {} & { $\alpha_2$ }                \\ \cline{2-2}\cline{4-7}\cline{9-9}
														   {} & {	\textit{$\delta_3$}	} & {} & { $e_{31}$ } & { $e_{32}$ } & { $\dots$  } & { $e_{3T}$ } & {} & { $\alpha_3$ }                \\ \cline{2-2}\cline{4-7}\cline{9-9}
														   {} & {	         \vdots  	} & {} & { $\vdots$ } & { $\vdots$ } & { $\ddots$ } & { $\vdots$ } & {} & { $\vdots                      $ }                \\ \cline{2-2}\cline{4-7}\cline{9-9}
														   {} & {	\textit{$\delta_S$}	} & {} & { $e_{S1}$ } & { $e_{S2}$ } & { $\dots$  } & { $e_{ST}$ } & {} & { $\alpha_S$ }                \\ 
					\cline{1-2}\cline{4-7}\cline{9-9}
					\noalign{\smallskip}
					\cline{1-2}\cline{4-7}\cline{9-9}
					\multirow{1}{*}{\centering\begin{turn}{90}V\end{turn}                     } & {        \textit{$\beta$   }	} & {} & { $\beta_1$ } & { $\beta_2$ } & {\dots} & { $\beta_T$ } & {} & { $0$ } \\ 
					\cline{1-2}\cline{4-7}\cline{9-9}
				\end{tabular}
			\end{center}
			\vspace{-7mm}
		\end{table}
	
	If a detection $\delta_s$ has been associated to a finger $\xi_t$, we have to define correspondences between the set of visible vertices of $\xi_t$ and the point cloud of the detection $\delta_s$. If the finger is already very close and the distance is below 10 mm, we do not compute any correspondences since the localization accuracy of the detector is not higher. In this case the finger is anyway close enough to the data to achieve a good alignment. 
	Otherwise, we compute the closest points between the vertices $\vertex_i$ and the points $\xClPoint_i$ of the detection:    
	\begin{equation}
	 E_{salient}(\para,D) = \sum_{s,t}e_{st} \left\{ \sum_{(\xClPoint_i,\vertex_i) \in \delta_s \times \xi_t} \Vert \vertex_i(\theta) - \xClPoint_i \Vert^2 \right\}
	\end{equation}
	As in \eqref{eq:errorResidual_p2pl}, a \emph{point-to-plane} distance metric can replace the \emph{point-to-point} metric. 
	When the overlap of the fingertip and the detection is less than 50\%, replacing the closest points $\xClPoint_i$ by the centroid of the detection leads to a speed up.

\vspace{-1mm}
\section{Experimental Evaluation}\label{sec:experimentalEvaluation}
\vspace{-2mm}

	Benchmarking in the context of 3D hand tracking remains an open problem \cite{Review_Erol_HandPose} 
	despite recent contributions \cite{srinath_iccv2013,GCPR_2013_Tzionas_Gall}. 
	Related RGB-D methods \cite{OikonomidisBMVC} usually report quantitative results only on synthetic sequences, which inherently include ground-truth, while for realistic conditions they resort to qualitative results.
	
	Although qualitative results are informative, 
	quantitative evaluation based on ground-truth is of high importance. 
	We therefore manually annotate $14$ new sequences,  
	$11$ of which are used to evaluate the components of our pipeline and $3$ 
	for comparison with the state-of-the-art method \cite{OikonomidisBMVC} (details in the supplementary material). 
	The standard deviation for $4$ annotators is $1.46$ pixels. 

	The error metric for our experiments is the 2D distance (pixels) between the projection of the 3D joints and the corresponding 2D annotations. 
	Details regarding the joints taken into account in the metric are included in the supplementary material. 
	We report the average error over all frames of all sequences.

	\vspace{-2mm}
	\subsection{Pipeline Components}\label{sec:evaluation_PipelineComponents}

	\begin{table}[t]	
		\footnotesize 
		\begin{center}
			\caption{	Evaluation of collision weights $\gamma_c$, using a 2D distance error metric (px). 
					Weight $0$ corresponds to deactivated collision detection, noted as ``$LO+S$'' in Table~\ref{table:switching_ON_OFF}. 
					Sequences are grouped (see supplementary material) in $3$ categories: 
					``\emph{Severe}''	 for intense, 
					``\emph{some}''		 for light and 
					``\emph{no apparent}''	 for imperceptible collision. 
					Our highlighted decision is based on the union of the ``\emph{severe}'' and ``\emph{some}'' sets, noted as ``\emph{at least some}'' 
				}
			\vspace{-4mm}
			\label{table:evaluationCollisionWeights}
			\setlength{\tabcolsep}{1pt}	
			\begin{tabular}{|c|c|c|c|c|c|c|c|c|c|}
				\cline{1-1}\cline{3-10}
				\multicolumn{1}{|c|}{$\gamma_c$} & \multicolumn{1}{c}{} & \multicolumn{1}{|c|}{$0$} & \multicolumn{1}{c|}{$1$} & \multicolumn{1}{c|}{$2$} & \multicolumn{1}{c|}{$3$} & \multicolumn{1}{c|}{$5$} & \multicolumn{1}{c}{$7.5$} & \multicolumn{1}{|c|}{$10$} & \multicolumn{1}{c|}{$12.5$} \\
				\cline{1-1}\cline{3-10}
				\noalign{\smallskip}
				\cline{1-1}\cline{3-10}
								
				\multirow{1}{*}{\textit{All}} 			& {} & {~$5.40$~} & {~$5.50$~} & {~$5.63$~} & {~$5.23$~} & {~$5.18$~} & {~$5.19$~} & 				{~$5.18$~} & {~$5.21$~}		\\ \cline{1-1}	\cline{3-10}
				\multirow{1}{*}{\textit{Only Severe}} 		& {} & { $6.00$ } & { $6.18$ } & { $6.37$ } & { $5.73$ } & { $5.66$ } & { $5.67$ } & 				{ $5.65$ } & { $5.71$ }		\\ \cline{1-1}	\cline{3-10}
				\multirow{1}{*}{\textit{Only Some}} 		& {} & { $3.99$ } & { $3.98$ } & { $3.98$ } & { $3.98$ } & { $3.98$ } & { $3.99$ } & 				{ $3.99$ } & { $3.98$ }		\\ \cline{1-1}	\cline{3-10}\hhline{-|~|--------}
				\multirow{1}{*}{\textit{~At least some~}} 	& {} & { $5.52$ } & { $5.65$ } & { $5.80$ } & { $5.31$ } & { $5.26$ } & { $5.27$ } & \cellcolor[gray]{0.8}	{ $5.25$ } & { $5.29$ }		\\ \cline{1-1}	\cline{3-10}
				
			\end{tabular}
		\end{center}
		\vspace{-4mm}
	\end{table}

	\begin{table}[t]	
		\footnotesize 
		\begin{center}
			\caption{	Evaluation of the parameter $\lambda$ of the assignment graph of Equation \eqref{eq:bipartite_MIP}, using a 2D distance error metric (px). 
					Value $\lambda=0$ corresponds to deactivation of the  detector, noted as ``$LO+C$'' in Table \ref{table:switching_ON_OFF}. 
					Both versions of $w_{s}$ described in Section \ref{subsec:salientPointDetector} are evaluated
			}
		\vspace{-4mm}
			\label{table:bipartiteTable_VIRTUAL_mean}
			\setlength{\tabcolsep}{1pt}	
			\begin{tabular}{|c|c|c|c|c|c|c|c|c|}
				\cline{3-9}
				\cline{1-1}\cline{3-9}
				\multicolumn{1}{|c|}{$\lambda$} & \multicolumn{1}{c}{} & \multicolumn{1}{|c}{\textit{0}} & \multicolumn{1}{|c|}{$0.3$} & $0.6$ & $0.9$ & $1.2$ & $1.5$ & $1.8$ \\
				\cline{1-1}\cline{3-9}
				\noalign{\smallskip}
				\cline{1-1}\cline{3-9}
				
				{ $w_{s}=1$}      			& {} & \multirow{2}{*}{\centering~$5.24$~} 	& {~$5.23$~} & {~$5.21$~} & {~$5.21$~} &                      {~$5.18$~} & {~$5.19$~} & {~$5.30$~} \\ \hhline{-|~||~|------}  
				{~$w_{s}=\frac{c_{s}}{c_{thr}}$~}   	& {} & 					  {} 	& { $5.21$ } & { $5.19$ } & { $5.18$ } & \cellcolor[gray]{0.8}{ $5.18$ } & { $5.28$ } & { $5.68$ } \\ \cline{1-1}\cline{3-9}
				
			\end{tabular}
		\end{center}
		\vspace{-4mm}
	\end{table}

	\begin{table}[t]	
		\footnotesize 
		\begin{center}
			\caption{	Evaluation of \emph{point-to-point} ($p2p$) and \emph{point-to-plane} ($p2plane$) distance metrics, along with iterations number of the optimization framework, 
					using a 2D distance error metric (px)
				}
			\vspace{-5mm}
			\label{table:evaluation_p2p_p2plane}
			\setlength{\tabcolsep}{1pt}	
			\begin{tabular}{|c|c|c|c|c|c|c|}
				\cline{1-1}\cline{3-7}
				\multicolumn{1}{|c|}{\textit{~Iterations~~}} & \multicolumn{1}{c}{} & \multicolumn{1}{|c|}{$5$} & \multicolumn{1}{c|}{$10$} & \multicolumn{1}{c|}{$15$} & \multicolumn{1}{c|}{$20$} & \multicolumn{1}{c|}{$30$} \\
				\cline{1-1}\cline{3-7}
				\noalign{\smallskip}
				\cline{1-1}\cline{3-7}
				
				\multirow{1}{*}{ $p2p$} 	& {} & {~$7.39$~} & 				{~$5.31$~} & {~$5.11$~} & {~$5.04$~} & {~$4.97$~}	\\ \cline{1-1}	\cline{3-7}\hhline{-|~|-----}
				\multirow{1}{*}{~$p2plane$~} 	& {} & { $5.39$ } & \cellcolor[gray]{0.8}	{ $5.18$ } & { $5.14$ } & { $5.13$ } & { $5.11$ }	\\ \cline{1-1}	\cline{3-7}
				
			\end{tabular}
		\end{center}
		\vspace{-5mm}
	\end{table}

	\begin{table}[t]	
		\footnotesize 
		\begin{center}
			\caption{	Evaluation of the components of our pipeline. ``$LO$'' stands for local optimization and includes fitting both \emph{data-to-model} ($d2m$) and \emph{model-to-data} ($m2d$), unless otherwise specified. 
					Collision detection is noted as ``$C$'', while salient point detector is noted as ``$S$''. 
					The number of sequences where the optimization framework collapses is noted in the last row, while the mean error is reported only for the rest
			}
			\vspace{-1mm}
			\label{table:switching_ON_OFF}
			\setlength{\tabcolsep}{1pt}	
			\begin{tabular}{|c|c|c|c|c|c|c|c|c|}
					   \cline{3-4}\cline{6-9}
				\cline{1-1}\cline{3-4}\cline{6-9}
				\multicolumn{1}{|c|}{\textit{Components}} & \multicolumn{1}{c}{} & \multicolumn{1}{|c}{~~$LO_{m2d}$~~} & \multicolumn{1}{|c|}{~$LO_{d2m}$~~} & \multicolumn{1}{c}{} & \multicolumn{1}{|c|}{~~~~$LO$~~~~} & ~~$LO+C$~~ & ~~$LO+S$~~ & ~$LO+CS$~ \\
				\cline{1-1}\cline{3-4}\cline{6-9}
				\noalign{\smallskip}
				\hhline{-|~|--|~|----}  
				
				{\textit{Mean Error (px)}}			& 		     {}	& 		     { $15.19$ } 	&	{ $   -   $ } 		& 		     {} 	& 			    	{ $5.59$ } & { $5.24$ } & { $5.40$ } & \cellcolor[gray]{0.8}	{ $5.18$ }	\\ \cline{1-1}\cline{3-4}\cline{6-9}
				{\textit{~Improvement (\%)~}  }	& \multicolumn{1}{c}{} 	& \multicolumn{1}{c}{} 			&	\multicolumn{1}{c}{} 	& \multicolumn{1}{c|}{} 		& \multicolumn{1}{c|}	{ $   -   $ } & { $6.39$ } & { $3.40$ } & 			{ $7.36$ }						\\ 
				\cline{1-1}
				\cline{6-9}
				\noalign{\smallskip}
				\cline{1-1}\cline{3-4}\cline{6-9}
			        {\textit{Failed Sequences}}	& {} & { $1/11$ } & { $11/11$ } & {} & { $  0/11 $ } & { $  0/11 $ } & { $ 0/11 $ } & 			{ $ 0/11 $ }\\ \cline{1-1}\cline{3-4}\cline{6-9}
				\cline{1-1}\cline{3-4}\cline{6-9}
			\end{tabular}
		\end{center}
		\vspace{-4mm}
	\end{table}

	Our system is based on an objective function consisting of four terms, described in Section \ref{sec:optimization}. 
	Two of them minimize the error between the posed mesh and the depth data by fitting the \emph{model to the data} and the \emph{data to the model}. 
	A \emph{salient point} detector further constraints the pose using fingertip detections in the depth image, 
	while a \emph{collision detection} method contributes to realistic pose estimates that are physically plausible.
	The above terms participate in the objective function in a weighted scheme, which is minimized as in~\cite{PonsModelBased}. 
	
	In the following, we evaluate the parameters used in our components and assess  
	both each component's individual contribution to the overall system performance, as well as of the combination thereof in the objective function \eqref{eq:obj}. 
	
	\vspace{-3mm}
	\subsubsection{Collision Detection}\label{subsubsec:experimentCollisionDetection}
	The collision detection component is regulated in the objective function \eqref{eq:obj} by the weight $\gamma_c$, so that collision and penetration get efficiently penalized. 
	Table \ref{table:evaluationCollisionWeights} summarizes our evaluation experiments for the values of $\gamma_c$. 
	Although we choose $\gamma_c=10$ for the present dataset, a generally proposed range of values for new sequences would be between $5$ and $10$. 
	
	\vspace{-3mm}
	\subsubsection{Salient Point Detection}\label{subsubsec:experimentSalientPointDetector}
	The salient point detection component 
	depends on the parameters $w_{s}$ and $\lambda$, as described in Section \ref{subsec:salientPointDetector}. 
	Table \ref{table:bipartiteTable_VIRTUAL_mean} summarizes our evaluation of the parameter $\lambda$ spanning
	a range of possible values for both cases $w_{s}=1$ and $w_{s}=\frac{c_{s}}{c_{thr}}$. 
	Although the difference between the two versions of $w_{s}$ is not very large, 
	the latter performs better for a wide range ($0.6$ to $1.2$) of parameter $\lambda$. 
	We therefore choose $w_{s}=\frac{c_{s}}{c_{thr}}$ and $\lambda=1.2$. 
	
	\vspace{-3mm}
	\subsubsection{Distance Metrics}\label{subsubsec:experimentDistanceMetrics}
	Table \ref{table:evaluation_p2p_p2plane} presents an evaluation of the two distance metrics presented in Section \ref{subsec:m2d}, 
	namely \emph{point-to-point} (Equation \eqref{eq:errorResidual_p2p}) 
	and \emph{point-to-plane} (Equation \eqref{eq:errorResidual_p2pl}), 
	along with the number of iterations of the minimization framework.
	The \emph{point-to-plane} metric leads to adequate minimization with only $10$ iterations, 
	providing a significant speed gain, being thus our choice. 
	However, we perform $50$ iterations for the first frame in order to ensure an accurate refinement of the manually initialized pose. 
	The runtime for the chosen setup (see supplementary material for benchmark details) is $2.74$ and $4.35$ 
	seconds per frame 
	for scenes containing one and two hands respectively. 
	
	\vspace{-3mm}
	\subsubsection{Component Evaluation}\label{subsubsec:experimentComponentEvaluation}
	Table \ref{table:switching_ON_OFF} presents the evaluation of each component and 
	the combination thereof. 
	Simplified versions of the pipeline, fitting either just the \emph{model to the data} or the \emph{data to the model} can lead to a 
	collapse of the pose estimation, due to unconstrained optimization.
	Our experiments quantitatively show the notable contribution 
	of both the collision detection and the salient point detector components.
	The best overall system performance is achieved with the combinatorial setup described by the objective function \eqref{eq:obj}. 
	Qualitative results are depicted in Figure \ref{fig:resultsQualitative}.

	\vspace{-3mm}
	\subsection{Comparison to State-of-the-Art}\label{sec:evaluation_FORTH}
	
\begin{figure}[b]
	\captionsetup[subfigure]{labelformat=empty}
	\centering
		\subfloat[subfigure 1 CompFORTH][]{
			\includegraphics[width=0.15 \textwidth]{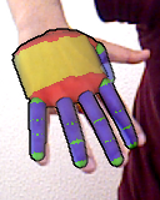}
			\label{fig:comp202forth}
		}
		\subfloat[subfigure 2 CompFORTH][]{
			\includegraphics[width=0.15 \textwidth]{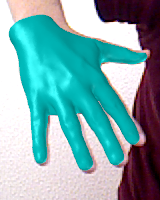}
			\label{fig:comp202mine}
		}
		\subfloat[subfigure 3 CompFORTH][]{
			\includegraphics[width=0.15 \textwidth]{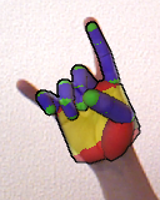}
			\label{fig:comp204forth}
		}
		\subfloat[subfigure 4 CompFORTH][]{
			\includegraphics[width=0.15 \textwidth]{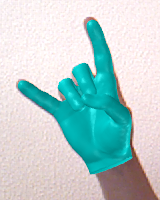}
			\label{fig:comp204mine}
		}
		\subfloat[subfigure 5 CompFORTH][]{
			\includegraphics[width=0.15 \textwidth]{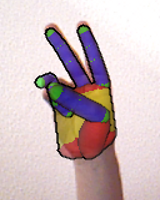}
			\label{fig:comp210forth}
		}
		\subfloat[subfigure 6 CompFORTH][]{
			\includegraphics[width=0.15 \textwidth]{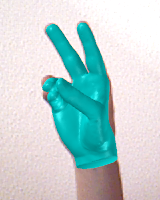}
			\label{fig:comp210mine}
		}
	\vspace{-7mm}
	\caption{	Qualitative comparison with \cite{OikonomidisBMVC}.
			Each image pair corresponds to the pose estimate of the FORTH tracker (left) and our tracker (right)
	}
	\label{fig:comparisonFORTH}
	\end{figure}

	Recently, Oikonomidis \etal~used particle swarm optimization (PSO) for a real-time hand tracker \cite{OikonomidisBMVC,Oikonomidis_1hand_object,Oikonomidis_2hands}. 
	These works constitute the state-of-the-art for single-view RGB-D hand tracking. 
	For comparison we use the software released for tracking one hand \cite{OikonomidisBMVC}, with the parameter setups of all the above works.
	Each setup is evaluated $3$ times 
	in order to compensate for the manual initialization and the inherent randomness of PSO. 
	Qualitative results depict the best version, while quantitative results report the average. 
	Figure \ref{fig:comparisonFORTH} qualitatively showcases the increased accuracy of our method, 
	along with the decreased accuracy of the FORTH tracker due to the sampling nature of PSO. 
	Quantitative results of Table \ref{table:ComparisonFORTH} show that our system outperforms \cite{OikonomidisBMVC} in terms of tracking accuracy. 
	However, it should be noted that the GPU implementation of \cite{OikonomidisBMVC} is real time, in contrast to our CPU implementation. 
	Detailed results for each evaluation of the parameter setups are included in the supplementary material.

	\begin{table}[t]	
		\footnotesize 
		\begin{center}
			\caption{	Comparison of our method 
					against the FORTH tracker. 
					We evaluate the FORTH tracker with $4$ parameter-setups 
					met in the referenced literature of the last column
				}
			\label{table:ComparisonFORTH}
			\setlength{\tabcolsep}{1pt}
			\vspace{-5mm}
			\begin{tabular}{|c|c|c|c|c|c|c|c|c|c|c|c|}
				
				\cline{5-7}\cline{9-10}\cline{12-12}
				\multicolumn{1}{c}{} & \multicolumn{1}{c}{} & \multicolumn{1}{c}{} & \multicolumn{1}{c|}{} & \textit{~Mean (px)~} & \textit{~St.Dev (px)~} & \multicolumn{1}{c|}{\textit{~Max (px)~}} & \multicolumn{1}{c}{} & \multicolumn{1}{|c|}{\textit{~Generations~}} & \multicolumn{1}{c|}{\textit{~Particles~}} & \multicolumn{1}{c}{} & \multicolumn{1}{|c|}{\textit{~Reference~}} \\
				\cline{5-7}\cline{9-10}\cline{12-12}
				\noalign{\smallskip}
				\cline{1-1}\cline{3-3}\cline{5-7}\cline{9-10}\cline{12-12}
				
				\multirow{4}{*}{\centering\begin{turn}{90}\textit{FORTH}\end{turn} 	 } & {} & \textit{~set 1~}	& {} & {$8.58$} & {$5.74$} & {$61.81$} & {} & {$25$} & {$64$}  & {} & {\cite{OikonomidisBMVC}} 			\\ \cline{3-3}	\cline{5-7}\cline{9-10}\cline{12-12}
													{} & {} & \textit{~set 2~}	& {} & {$8.32$} & {$5.42$} & {$57.97$} & {} & {$40$} & {$64$}  & {} & {\cite{Oikonomidis_1hand_object}} 		\\ \cline{3-3}	\cline{5-7}\cline{9-10}\cline{12-12}
													{} & {} & \textit{~set 3~}	& {} & {$8.09$} & {$5.00$} & {$38.90$} & {} & {$40$} & {$128$} & {} & {\cite{LucaHands}} 				\\ \cline{3-3}	\cline{5-7}\cline{9-10}\cline{12-12}
													{} & {} & \textit{~set 4~}	& {} & {$8.16$} & {$5.18$} & {$39.85$} & {} & {$45$} & {$64$}  & {} & {\cite{Oikonomidis_2hands}} 			\\ 		\cline{5-7}\cline{9-10}\cline{12-12}
				\cline{1-1}\cline{3-3}\cline{5-7}\cline{9-10}\cline{12-12}
				\noalign{\smallskip}
				\cline{1-3}\cline{5-7}\hhline{---|~|---}  
				\multicolumn{3}{|c|}{\textit{Proposed} } 		      			   & {} & \cellcolor[gray]{0.8}{ $3.76$ } & {$2.22$} & { $19.92$ } & \multicolumn{1}{c}{} & \multicolumn{1}{c}{} & \multicolumn{1}{c}{} & \multicolumn{1}{c}{} & \multicolumn{1}{c}{} 	\\ \cline{3-3}
				\cline{1-3}\cline{5-7}
				
			\end{tabular}
			\vspace{-6mm}
		\end{center}
	\end{table}

\section{Conclusion}\label{sec:conclusion}
\vspace{-2mm}
	
	In this work we have presented a system capturing the motion of two highly interacting hands. 	
	Inspired by the recent method~\cite{LucaHands}, we propose a combination of a generative model with a discriminatively trained salient point detector and collision modeling to obtain accurate, realistic pose estimates with increased immunity to ambiguities and tracking errors. While~\cite{LucaHands}  
	depends on expensive, specialized equipment, we achieve accurate tracking results using only a single cheap, off-the-shelf RGB-D camera. We have evaluated our approach on 14 new challenging sequences and shown that our approach achieves a better accuracy than the state-of-the-art single hand tracker~\cite{OikonomidisBMVC}.

\section{Acknowledgments}\label{sec:acknowledgements}
\vspace{-3mm}
	The authors acknowledge 
	the help of Javier Romero and Jessica Purmort of MPI-IS regarding the acquisition of the personalized hand model, 
	the assistance of Philipp Rybalov with annotation and 
	the public software release of the FORTH tracker by the CVRL lab of FORTH-ICS, enabling comparison to \cite{OikonomidisBMVC}. 
	Financial support was provided by the DFG Emmy Noether program (GA 1927/1-1).

\vspace{-3mm}
	
\newpage
\bibliographystyle{splncs03}
\bibliography{egbib}

\end{document}